\documentclass[runningheads]{llncs}
\usepackage[T1]{fontenc}
\usepackage{graphicx}
\usepackage{booktabs}
\usepackage{tabularx}
\usepackage{multirow}
\usepackage{amssymb}
\usepackage{makecell}
\usepackage{enumitem}
\usepackage{longtable, multirow, booktabs}
\usepackage{xcolor}
\usepackage{geometry}
\usepackage{fontspec}

\newcommand{\FrameworkName}{N2N}

\makeatletter
\def\ps@headings{%
  \def\@oddhead{\null}
  \def\@evenhead{\null}
  \def\@oddfoot{\null\hfill\thepage\hfill\null}
  \def\@evenfoot{\null\hfill\thepage\hfill\null}
}
\makeatother
\begin{document}
\title{\FrameworkName{}: A Parallel Framework for Large-Scale MILP under Distributed Memory}
\titlerunning{}
%
\author{
Longfei Wang\inst{1} 
\and Junyan Liu\inst{2} 
\and Fan Zhang\inst{2} 
\and Jiangwen Wei\inst{1} 
\and Yuanhua Tang\inst{1} 
\and Jie Sun\inst{2} 
\and \\ Xiaodong Luo\inst{1, 3} \thanks{Corresponding author.}
}

\authorrunning{}
%
\institute{
Shenzhen Research Institute of Big Data \\
\email{
\{wanglongfei,weijiangwen,tangyuanhua,xdluo\}@sribd.cn}
\and Theory Lab, Central Research Institute, 2012 Labs, Huawei Technologies Co. Ltd., Hong Kong SAR, China \\
\email{\{liu.junyan1,zhang.fan2,j.sun\}@huawei.com}
\and Chinese University of Hong Kong, Shenzhen
}
\maketitle              
\begin{abstract}
Parallelization has emerged as a promising approach for accelerating MILP solving. However, the complexity of the branch-and-bound (B\&B) framework and the numerous effective algorithm components in MILP solvers make it difficult to parallelize. In this study, a scalable parallel framework, \FrameworkName{} (a \emph{node-to-node} framework that maps the B\&B nodes to distributed computing nodes), was proposed to solve large-scale problems in a distributed memory computing environment. Both deterministic and nondeterministic modes are supported, and the framework is designed to be easily integrated with existing solvers. Regarding the deterministic mode, a novel sliding-window-based algorithm was designed and implemented to ensure that tasks are generated and solved in a deterministic order. Moreover, several advanced techniques, such as the utilization of CP search and general primal heuristics, have been developed to fully utilize distributed computing resources and capabilities of base solvers. Adaptive solving and data communication optimization were also investigated. A popular open-source MILP solver, SCIP, was integrated into \FrameworkName{} as the base solver, yielding \FrameworkName{}-SCIP. Extensive computational experiments were conducted to evaluate the performance of \FrameworkName{}-SCIP compared to ParaSCIP, which is a state-of-the-art distributed parallel MILP solver under the UG framework. The nondeterministic \FrameworkName{}-SCIP achieves speedups of 22.52 and 12.71 with 1,000 MPI processes on the Kunpeng and x86 computing clusters, which is 1.98 and 2.08 times faster than ParaSCIP, respectively. In the deterministic mode, \FrameworkName{}-SCIP also shows significant performance improvements over ParaSCIP across different process numbers and computing clusters. To validate the generality of \FrameworkName{}, HiGHS, another open-source solver, was integrated into \FrameworkName{}. The related results are analyzed, and the requirements of \FrameworkName{} on base solvers are also concluded.

\end{abstract}
\section{Introduction}

Mixed-integer linear programming (MILP), in the following form, has served as a powerful decision-optimization methodology in planning, scheduling, routing, and resource allocation since the 1950s~\cite{junger200950}. 

\begin{equation}
\min\limits_{x}\{\langle c,x \rangle : Ax \leq b, x \in \mathbb{R}^{n}, x_{I} \in \mathbb{Z}^{|I|}\}
\label{milp}
\end{equation}

where $A \in \mathbb{R}^{m\times n}, b \in \mathbb{R}^{m}, c \in \mathbb{R}^{n}, I \subseteq \{1,\ldots,n\}$.

Numerous algorithms and software~\cite{achterberg2009scip} have been investigated and developed to solve MILP problems efficiently, leading to general-purpose solvers capable of finding optimal solutions with high efficiency (such as OptVerse~\cite{optverse}, Gurobi~\cite{gurobi}, COPT~\cite{copt}, and FICO Xpress~\cite{xpress}). However, solving large-scale MILP problems in real-world applications within a reasonable time remains challenging. Therefore, theories and technologies from other fields, including artificial intelligence and multi-machine and multi-core hardware-based parallel computing, are integrated to accelerate the solving of large-scale MILP problems. The integration of AI and optimization, or learning to optimize, has been intensively investigated in recent years, and excellent achievements have been obtained. Reference~\cite{huang2025milp} thoroughly explored learning-based technologies in branch-and-bound, and potential directions and extensions to further enhance the efficiency of solving different MILP problems were proposed in the study. In~\cite{huang2022improving}, a Bi-layer Prediction-based Reduction Branch (BP-RB) framework was proposed to speed up the process of finding a high-quality feasible solution. In this framework, a graph convolutional network (GCN) is employed to predict the values of binary variables, which are used to fix a subset of binary variables. Moreover, the second layer GCN is employed to update the prediction of the values of the remaining binary variables and improve the branching algorithms in B\&B. Numerical experiments show that the framework can significantly accelerate primal heuristics. Other valuable studies have also been proposed to use learning-based method to improve the algorithm components in B\&B, including presolving~\cite{liu2024l2p}, branching~\cite{khalil2016learning}, cut selection~\cite{huang2022learning}, node selection~\cite{zhang2023reinforcement}, and symmetry handling~\cite{chen2024symilo}.

Meanwhile, rapid advancements in both software and hardware technologies have enabled effective parallelization across a wide range of domains, thereby providing new opportunities for accelerating the solving of MILPs. Branch-and-bound (B\&B), the most fundamental and important solving framework in state-of-the-art MILP solvers, follows a sequence of procedures that closely parallels those of conventional tree-search algorithms~\cite{land2009automatic}. This procedure suggests an intuitive parallelization approach: maintaining a B\&B tree centrally and distributing nodes to processors for solving in parallel to reduce the overall solving time. Over the past three decades, both commercial solvers and open-source projects have investigated the parallel implementations of MILP solvers. In this study, we focused on MILP solving acceleration through multi-machine and multi-core parallel computing, and novel algorithms and techniques were developed.

\subsection{Related Literature and Software}
Existing parallel implementations of MILP solvers can be classified into two categories: (1) parallel solvers, in which the parallel algorithms are tightly integrated with the underlying sequential procedures~\cite{berthold2018parallelization,cbc,ralphs2005symphony}, and (2) parallel frameworks, which are generic frameworks that aim to parallelize base solvers from ``outside''~\cite{shinano2018ubiquity,xu2005alps}. Because parallelization from the ``outside'' inevitably incurs overhead due to data transformation and communication, and because certain internal data structures and operations of base solvers cannot be accessed or controlled by external frameworks, parallel solvers may offer superior performance compared with parallel frameworks. Nevertheless, parallel frameworks require considerably less development effort and can be more easily integrated with diverse existing solvers.

To provide a coherent overview of prior work, several key concepts and features must first be clarified. Parallel granularity is commonly categorized into four levels: tree, subtree, node, and subnode parallelism. Parallel paradigms typically include the supervisor–worker (or scheduler–worker, S–W), master–worker (M–W), and master–hub–worker (M–H–W) models. Communication architectures can further be distinguished as either distributed memory (DM) or shared memory (SM). Building upon these definitions, representative implementations are summarized in Table \ref{tab:para_solver_framework}.

\begin{table}[htbp]
    \caption{Parallel Implementations of MILP Solvers}
    \label{tab:para_solver_framework}
    \begin{tabularx}{\textwidth}{X X X X X X}
        \toprule
        Solver/Framework & Year & Organization & Paradigm & Granularity & Architecture \\
        \midrule
        
        FiberSCIP~\cite{shinano2018fiberscip} & 2013 & ZIB & S-W & Subtree & SM \\
        ParaSCIP~\cite{shinano2016solving,shinano2011parascip} & 2011 & ZIB & S-W & Subtree & DM \\
        
        ParaXpress~\cite{shinano2016first} & 2016 & ZIB & S-W & Subtree & DM \\
        ParaLEX~\cite{shinano2007paralex,shinano2008dynamic} & 2007 & TUAT & M-W & Subtree & DM \\
        
        PUBB2~\cite{shinano2003effectiveness} & 2003 & TUAT & M-W & Subtree & DM \\
        
        PartiMIP~\cite{lin2025parallel} & 2025 & CAS & S-W & Subtree & SM \\
        
        PEBBL~\cite{eckstein2015pebbl} & 2015 & RU & M-H-W & Subtree & DM \\
        
        CBC~\cite{cbc} & $<$2015 & COIN-OR & M-W & Subtree & SM \\
        
        CHiPPS~\cite{xu2009computational} & 2009 & COIN-OR & M-H-W & Subtree & DM/SM \\
        
        SYMPHONY~\cite{ralphs2005symphony} & 2005 & COIN-OR & M-W & Node & DM/SM \\
        
        DIP~\cite{galati2012computational} & 2012 & COIN-OR & M-H-W & Subtree & SM \\
        
        FICO Xpress~\cite{berthold2018parallelization} & 2018 & FICO & S-W & Subtree & SM \\
        
        FATCOP ~\cite{chen2001fatcop} & 2001 & UW–Madison & M-W & Node & DM \\
        
        ZRAM~\cite{marzetta1998zram,brungger1999parallel} & 1998 & ETH Z\"urich & M-W & Subtree & DM \\
        \bottomrule
    \end{tabularx}
\end{table}

The basic idea of parallelizing B\&B is to distribute the tree nodes to the processors for parallel processing. Because only the root node exists when the solving process begins, the parallel solving process can be split into three phases in a natural way: ramp-up, primary, and ramp-down. Two types of ramp-up mechanisms have been proposed in the UG framework~\cite{shinano2018fiberscip}: normal and racing ramp-up. In the normal ramp-up, the workers that already received open nodes solve the corresponding subproblem and send half of the child nodes to the supervisor (which is named Load Coordinator in the UG framework), and the supervisor assigns these nodes to idle workers. In the racing ramp-up, all workers start solving the root node using different parameter settings, and a winner is selected based on certain criteria. The left open nodes of the winner are collected by the supervisor, and other workers are requested to terminate the solving. The collected nodes are then distributed to idle workers and the primary phase begins. Other ramp-up or initialization methods, including root/enumerative/selective/direct initialization, two-level root initialization, and spiral initialization~\cite{xu2009computational}, have been investigated in previous studies. However, an efficient ramp-up mechanism is still worth investigating, especially when the root node is time-consuming to process or the number of workers is large.

Load balancing is the most important factor affecting the computational performance in the primary phase. In the UG framework, several advanced techniques, such as the collecting mode and the delayed processing of nodes, have been developed to achieve highly efficient dynamic load balancing~\cite{shinano2018ubiquity,shinano2016solving,shinano2016first}. However, communication may become a bottleneck in large-scale clusters, and uneven workload distribution makes it difficult to fully utilize the available computing resources.

Another key issue for parallelization is determinism,  which is necessary for debugging, performance evaluation, and fulfilling users' requirements for reproducibility. In the FICO Xpress-Optimizer, a deterministic parallel scheme based on deterministic stamps and a trailing read barrier was developed, achieving a high level of CPU utilization~\cite{berthold2018parallelization}. In contrast, achieving an effective balance between determinism and parallel performance is more challenging in general-purpose parallel frameworks. As demonstrated in UG's token-based implementation~\cite{shinano2018fiberscip}, deterministic execution often yields inferior performance compared with sequential solvers.

As discussed above, although there have been many excellent research results, four issues or challenges still need to be further investigated, and this study focused on these issues:
\begin{enumerate}
    \item Deterministic parallel algorithm: How to design and improve data synchronization, knowledge sharing, and task scheduling in deterministic parallel solving?
    \item Utilization of parallel computing resources and capabilities of base solvers: How to generate and schedule nontrivial solving tasks to utilize and coordinate workers?
    \item Adaptive solving: How to improve the efficiency and robustness of parallel solvers through adaptive strategies?
    \item Data communication: How to minimize the data transmission overhead in large-scale distributed clusters? How to ensure data consistency between the supervisor/master and workers? 
\end{enumerate}

\subsection{Contributions} 
We propose a parallel framework based on the Supervisor–Worker paradigm, employing subtree-level parallelism for large-scale cluster environments. The main contributions of this work are as follows:

\begin{enumerate}
    \item Novel deterministic algorithm: We introduce a novel sliding-window–based algorithm that enhances knowledge sharing and enables flexible load balancing while strictly preserving determinism.
    \item Leveraging hardware and software capabilities: The utilization of parallel computing resources in large clusters will be limited if B\&B tree nodes are distributed and solved in a naive parallel method. For example, the shape of B\&B tree may be extremely unbalanced, and there are not enough nodes to be solved in parallel. Beyond B\&B, additional solving strategies—including primal heuristics and constraint programming—are incorporated during the ramp-up and primary phases. 
    \item Adaptive solving: Solver parameters and the set of active workers are adaptively adjusted according to problem characteristics and cluster-level resource information. Moreover, multiple presolving strategies are used to obtain presolved instances with smaller size.
    \item Highly optimized data communication: In \FrameworkName{}, the presolved instance is transferred in the MPS format, which guarantees consistency between the instances processed by the supervisor and workers. This method addresses the unreliable and complicated issues in the method of transferring model data in memory, which is used in the UG framework. Moreover, extensive optimizations were performed to minimize communication overhead in the parallel solving process in large-scale distributed settings.
    \item Compatibility with Huawei Kunpeng: To our knowledge, this is the first study that investigates the solving capability of parallel frameworks on both x86 and Huawei Kunpeng platforms. The numerical results further indicate that superior speedups can be achieved on Huawei Kunpeng clusters.
\end{enumerate}
 
The remainder of this paper is organized as follows. Section 2 introduces the overall architecture of \FrameworkName{} and its core algorithms. Some important implementation details are described in Section 3. Section 4 presents the performance evaluation of \FrameworkName{}-SCIP (implemented with SCIP as the base solver) and comparisons with ParaSCIP. Moreover, the computational performance of \FrameworkName{}-HiGHS (implemented with HiGHS as the base solver) was evaluated, and the results are discussed. Finally, we conclude with a summary of our contributions and future research directions.

\section{Architecture and Core Designs in \FrameworkName{}}

\begin{figure}[htbp]
    \centering
    \includegraphics[width=1.0\linewidth]{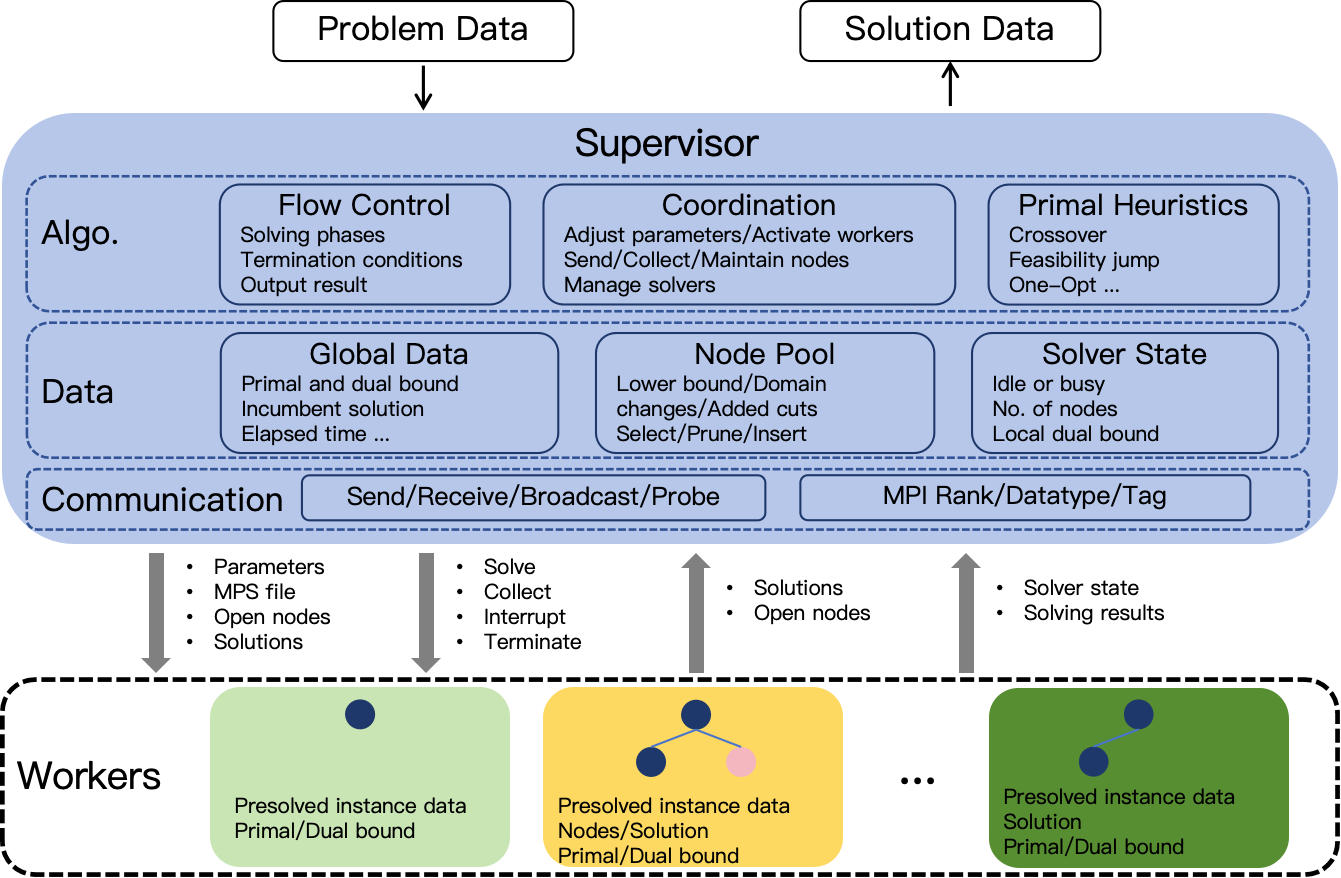}
    \caption{Architecture of \FrameworkName{}}
    \label{fig:architecture}
\end{figure}

The architecture of the \FrameworkName{} framework is illustrated in Fig. \ref{fig:architecture}. \FrameworkName{} employs a Supervisor-Worker paradigm to achieve subtree-level parallelism.

\FrameworkName{} extends the solving phases by adding a dedicated preprocessing phase before the ramp-up phase. In the preprocessing phase, the supervisor analyzes the problem structure and selects tailored solving strategies and parameters. The original problem is presolved by the supervisor, and the presolved problem is written into an MPS format file, which is sent to each machine in the computing cluster. To provide a clear description of the ideas in the \FrameworkName{} framework, the main differences between the \FrameworkName{} and UG frameworks are listed in Table~\ref{tab:diff_n2n_ug}. The innovative algorithms and techniques in \FrameworkName{} are summarized in Table~\ref{tab:summary_inno_tech}. Some techniques are discussed in detail in the following sections.

\begin{table}[htbp]
    \centering
    \caption{Comparisons Between \FrameworkName{} and UG}
    \label{tab:diff_n2n_ug}
    \begin{tabular}{l l l}
    \toprule
        Features & \FrameworkName{} & UG \\
        \midrule

        Determinism & \makecell[tl]{Sliding-window-based \\ algorithm} & Token-based algorithm \\

        \hline
        \makecell[tl]{Communication \\architecture} & Distributed memory & \makecell[tl]{Distributed memory and \\shared memory} \\

        \hline
        \makecell[tl]{Preprocessing \\ phase} & \makecell[tl]{Detect problem structure \\ Adjust parameters \\ Multi-Presolving} & \makecell[tl]{No dedicated \\ preprocessing phase} \\

        \hline
        \makecell[tl]{General \\ primal \\ heuristics} & \makecell[tl]{Crossover \\ Feasibility jump \\ One-Opt} & No heuristics \\

        \hline
        CP search & \makecell[tl]{Utilize CP search in SCIP} & No CP search \\        
    \bottomrule
    \end{tabular}

\end{table}

\begin{table}[htbp]
    \centering
    \caption{Summary of Techniques in \FrameworkName{}}
    \label{tab:summary_inno_tech}
    \begin{tabular}{l p{10cm}}
    \toprule
        Category & Algorithms/Techniques \\
        \midrule
        \makecell[tl]{Novel deterministic \\ algorithm} & A novel sliding-windows-based algorithm to ensure that tasks are generated and solved in a deterministic order \\

        \hline
        \multirow{3}{*}{\makecell[tl]{Leveraging \\ hardware and \\ software \\ capabilities}} & $\bullet$ Set some solvers to emphasize on CP-like search when integrating with SCIP to take full advantage of the CP techniques in SCIP \\
        & $\bullet$ Several general primal heuristics, including crossover, feasibility jump and one-opt, were developed in the framework \\
        & $\bullet$ A supplementary double presolving technique is developed in the framework to help base solvers solve subproblems more efficiently \\

        \hline
        \multirow{2}{*}{\makecell[tl]{Adaptive \\solving}} & $\bullet$ Adjust parameters and select workers to activate adaptively \\
        & $\bullet$ Multi-Presolving: Presolve the original instance with multiple strategies and select the smallest presolved instance \\

        \hline
        \multirow{2}{*}{\makecell[tl]{Highly optimized \\ data \\ communication}} & $\bullet$ Transferring presolved instances in MPS format file \\
        & $\bullet$ Packing data; Using the data type with the smallest size; Encoding data  \\
    \bottomrule
    \end{tabular}

\end{table}

\subsection{Novel Deterministic Algorithm}

A sliding-window-based deterministic algorithm was designed and implemented in the \FrameworkName{} framework to address two fundamental challenges: (1) maintaining strict reproducibility without sacrificing parallel efficiency, and (2) minimizing the synchronization overhead during parallel processing.

\begin{figure}[htbp]
    \centering
    \includegraphics[width=1.0\linewidth]{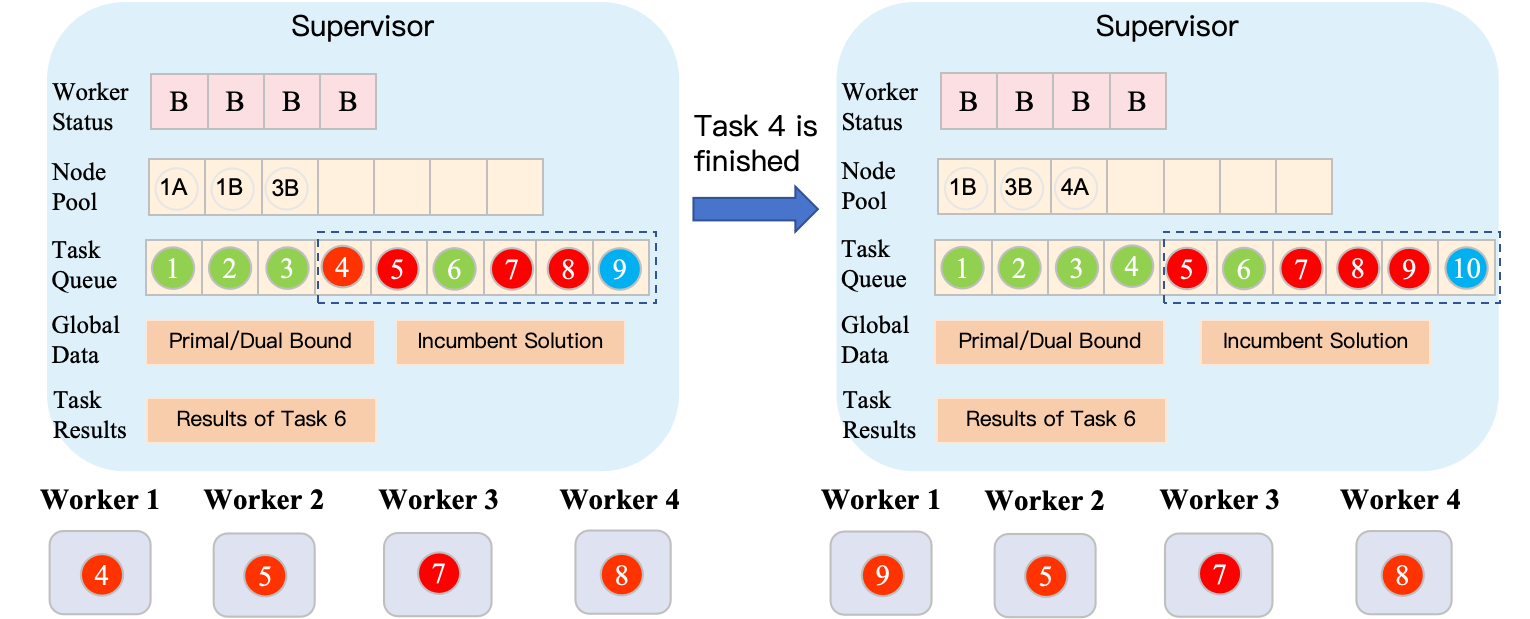}
    \caption{Illustration of the Sliding-Window-Based Algorithm}
    \label{fig:sliding_window}
\end{figure}

The sliding-window-based algorithm involves a key concept referred to as ``task''. When generating or defining a solving task, the corresponding subproblem (open node) should be given, and parameters (especially for the termination criteria) should be specified. Moreover, the primal bound and incumbent solution can also be included in the task to assist in processing the task. On the assumption that the base solver is deterministic, which is a trivial assumption for most mature solvers, if tasks are generated and solved in a deterministic order, the overall parallel algorithm is deterministic.

Let $W$ denote the width of the sliding-window. In the sliding-window-based algorithm, $W$ tasks are first generated and released based on the open nodes and global data (including the primal bound and incumbent solution). These open nodes and global data are collected after the ramp-up phase, during which the supervisor solves the root node and stops when the deterministic termination conditions are met. The parameters in each task are also deterministic. The tasks in the sliding-window can be solved independently by workers. If the task in the leftmost of the window is finished, then the window slides to the right by one unit, that is, one new task is generated and released based on the global data. For each solver, its state is reset after solving a task. Therefore, the solving results of each task are deterministic, regardless of which solver is selected to solve.

Fig. \ref{fig:sliding_window} is an illustration of the sliding-window-based algorithm. In the illustration, there are 4 workers and the width of the sliding-window is 6. ``B'' in the ``Worker Status'' indicates that the worker is busy. In the left part of the figure, Task 4, Task 5, Task 7 and Task 8 are being processed by the four workers simultaneously. There are two other tasks in the sliding-window (the dashed box). Task 6 has been finished, and its results has been sent to and stored in the supervisor, but will not be merged into the node pool and global data until the results of Task 5 are merged. Task 9 has been generated and is waiting to be processed. There are 3 open nodes in the node pool, and ``1A'' indicates that it is one of the left nodes of Task 1. After Task 4 is finished, the window slides to the right. As shown in the right part of the figure, after the results of Task 4 are synchronized to the node pool (node ``4A'') and global data, the first node (``1A'') in the node pool is popped and is used to generate the new Task 10. Additionally, because Task 4 has been finished by Worker 1, so Task 9 is assigned to it.

The sliding-window-based algorithm is easy to implement, and there is no explicit use of any lock, token or timer mechanisms. The tasks in the window can be solved independently by any available worker. Workers have no need for mutual waiting or inter-worker communication, enabling better utilization of parallel computing resources. Another advantage is that the determinism only depends on the width of the sliding-window, that is, the solving process and results remain the same even when the number of workers changes.

\subsection{Leveraging Hardware and Software Capabilities}
In the \FrameworkName{} framework, some advanced techniques have been developed to fully utilize parallel computing resources and capabilities of base solvers to achieve good solving performance.

\begin{enumerate}
    \item Constraint Programming Search: The powerful constraint programming search techniques in SCIP may be more effective in solving some types of problems than the B\&B tree search. In the preprocessing phase of the \FrameworkName{} framework, the supervisor identifies whether the problem is a CP problem (i.e., whether it has an objective function). If the problem is a CP problem, the supervisor attempts to solve it based on CP-like search strategies in SCIP in a short time. If the problem is solved successfully, the solving process is terminated. In the ramp-up phase of the nondeterministic mode, the racing ramp-up strategy is used, and some solvers are configured to emphasize the CP-like search. If any of these CP solvers successfully solves the problem, the solving process can be terminated.
    \item Primal Heuristics: Some general primal heuristics are developed in the parallel framework to improve the efficiency of obtaining good feasible solutions. One primal heuristic is crossover~\cite{berthold2006primal}, a classical and powerful primal heuristic that can be easily parallelized. When the crossover algorithm is triggered, good feasible solutions in the solution pool are used to generate a sub-MIP based on their common integer values. Each generated sub-MIP is then used to construct an open node, which is assigned to an idle solver to attempt the discovery of improved feasible solutions.

    Another primal heuristic is the feasibility jump~\cite{luteberget2023feasibility}, which was proposed in recent years and has a strong ability to find feasible solutions quickly. A general feasibility jump was implemented in the \FrameworkName{} framework and is invoked before a solver starts solving a subproblem. If a feasible solution is found, it is passed to the base solver as the starting solution. Moreover, One-Opt is used to further improve the quality of solutions obtained from the feasibility jump.

    \item Supplementary Double Presolving: In addition to the presolving for the original instance in the supervisor, workers presolve the received node again before solving it. This is called ``double presolving''. If ``double presolving'' is turned on, it is necessary to transform the data back to the original space (analogous to postsolving) before sending it to the supervisor. However, HiGHS lacks functions to map presolved data (e.g., domain modifications or added cuts) back to the original space. Therefore, presolving should be turned off when using the HiGHS solver to solve subproblems. To mitigate the resulting efficiency loss, the \FrameworkName{} framework implements a supplementary double presolving technique. This technique removes fixed variables from subproblems prior to solving and restores them when sending results back to the supervisor. By reducing the size of subproblems, this approach enhances HiGHS’s solving efficiency. Moreover, the technique is applicable to other base solvers if they lack the transformation functions.
\end{enumerate}

\subsection{Adaptive Solving}
In practice, machine memory is often limited and can be exhausted if too many processes run concurrently. To address this issue, the proposed framework adaptively selects the subset of workers to be activated for solving subproblems. During the preprocessing phase, the supervisor determines the number of active workers based on the characteristics of the presolved instance, memory information collected from each machine in the cluster, and user-specified configurations for MPI processes.

\subsection{Highly Optimized Data Communication}

To reduce the communication overhead and ensure the data consistency between computing nodes, some highly optimized data communication techniques have been developed in the \FrameworkName{} framework.

\begin{enumerate}
    \item Method for Transferring Presolved Instances: At the beginning of the parallel solving process, the original instance is presolved by the supervisor, then the presolved instance data is transferred to workers. In the ramp-up and primary phase, only the differences between the open nodes and the presolved instance, that is, the domain changes and added cuts, will be transferred between the supervisor and workers. Even though the presolved MILP model can be represented by several simple vectors and a constraint matrix, the data structures and internal transformations may be very complicated in base solvers. For example, the constraints in the presolved instances are transformed and classified into different types in the SCIP solver environment. Therefore, it is necessary to design a generic and safe method to transfer presolved instance data. In the \FrameworkName{} framework, the presolved instance is serialized into the MPS format and transmitted as a file across the cluster. Workers then ingest this file to initialize their respective solving environments. Given that the MPS format is the most prevalent, solver-independent standard for representing and storing MILP problem data, this file-based transfer method guarantees the generality and interoperability of the presolved instance data across diverse base solvers.
    \item MPI Data Transmission Optimization: In \FrameworkName{}, the MPI (Message Passing Interface) is used to transmit data in the distributed computing environment. Several techniques were developed to minimize the data transmission overhead, including: (1) Packing several small related data into a big data structure to decrease the number of data transmission events; (2) Using the data type with the smallest size to store and transfer data; (3) Encoding several data into one data field.
\end{enumerate}

\section{Implementation Details}
The framework was developed using C++14, and communication between processes in the distributed computing cluster was implemented based on the MPICH.

\subsection{Integration with Base Solvers}
\FrameworkName{} was designed as a general framework; therefore, an abstract design was implemented for common modules, including the supervisor, base solvers, and node pool. Integrating with the base solver yields a parallel MILP solver, which can be invoked through the following approach:

\begin{center}
    \verb|mpirun -np 1000 N2N-[BaseSolver] parameter_file.set instance_file.mps|
\end{center}

The foregoing command means that MPI environment is launched to run the parallel solver in the distributed cluster. ``\verb|-np 1000|'' indicates that the number of MPI processes is set to 1,000. ``\verb|N2N-[BaseSolver]|'' is the executable binary file, and \verb|parameter_file.set| is the file to store parameter settings, followed by the MPS file path for the instance to be solved.

Currently, SCIP (version 9.1.1) and HiGHS (version 1.11.0) have been integrated into the framework. When integrated with SCIP, the plugin interfaces in SCIP were used to implement algorithms and transfer data. When integrating with HiGHS, some callback functions were defined and set using interfaces of HiGHS. These plugins or callback functions can be classified into four types: (1) receive and process the interruption and termination request message from the supervisor; (2) transfer and send the new incumbent solution immediately; (3) send solver state and tree nodes to the supervisor; and (4) receive and process the primal bound and incumbent solutions from the supervisor.

\subsection{Data Acquisition from Base Solvers}
Detailed data from the base solvers are essential for the solving process. When integrated with SCIP, numerous well-designed APIs facilitate data acquisition and transformation. In contrast, HiGHS lacks such interfaces, particularly for accessing information on added cuts and domain changes at nodes. Consequently, these data are extracted directly from the data structures in HiGHS, such as HighsCutPool, HighsNodeQueue, and HighsDomainChange.

\section{Computational Results}

\subsection{Experimental Setup}

\subsubsection{Test Set}

An appropriate test set for evaluating parallel B\&B algorithms should satisfy two fundamental criteria~\cite{ralphs2018parallel}: (1) the search trees should contain sufficient nodes to evaluate the load-balancing capability of parallel implementations; and (2) all problems should be solvable by a single-threaded sequential solver to enable meaningful performance comparisons between the sequential and parallel approaches. Therefore, careful selection of a suitable test set is essential.

For the nondeterministic test set, we constructed an initial test set by combining the MIPLIB 2017 Collection, which contains 1,065 instances from broad applications, with 25 additional problems from the parallel MILP solving literature. The unsolved (open) problems were removed from the initial dataset. Subsequently, experiments were conducted on the remaining instances using single-threaded SCIP on nearly 20 workstations. Based on the experimental results, problems that took less than 1 hour or more than 24 hours to solve, and problems with fewer than 1,000 B\&B nodes, were eliminated. The test set was further refined by removing problems that failed to be solved or showed significant variations in the node number across multiple runs with different random seeds. The final test set contains 50 instances, and 24 hours is set as the time limit when testing.

For the deterministic mode, the performance of ParaSCIP with the deterministic mode on the selected 50 instances was first tested on a server. Unfortunately, the deterministic ParaSCIP failed to solve all 50 instances within the 24 hours time limit. For a reasonable test span and comparison, more appropriate instances should be selected to evaluate the deterministic algorithm. First, both single-threaded SCIP and 16-threads FiberSCIP on the MIPLIB 2017 Benchmark were tested: deterministic FiberSCIP successfully solved 100 out of 240 problems, among which 14 instances were solved faster than by single-threaded SCIP. Subsequently, 4 instances that were solved within 1 minute by the single-threaded SCIP were removed. The remaining 10 instances make up the final dataset for the deterministic mode. When testing with the deterministic mode, 7,200 seconds was set as the time limit.

\subsubsection{Solver Versions} In the computational experiments, SCIP and ParaSCIP from SCIP Optimization Suite (version 9.1.1) were built and tested. HiGHS (version 1.11.0) was integrated into \FrameworkName{} and tested. 

\subsubsection{Cluster Configurations}
Computational experiments were conducted on two clusters with different architectures: Huawei Kunpeng and x86. The Kunpeng cluster contains 8 servers, each of which is equipped with Kunpeng CPUs, which are designed based on the ARM architecture, and 256 GB of memory. The x86 cluster contains 10 servers, each of which is equipped with CPUs based on x86 architecture and 753 GB of memory.

\subsubsection{Metrics}
Three metrics are used for comparison: (1) the number of instances (\# solved) that are solved to optimality in the given time limit; (2) the shifted geometric mean (SGM, with a shifted value of 10.0 in this study) evaluating the average solving time; and (3) speedup measuring the efficiency of parallel algorithms relative to the corresponding sequential baseline.

\subsection{Computational Performance of \FrameworkName{}-SCIP}

\subsubsection{Performance with 1,000 Processes}

As shown in Table~\ref{tab:nondet_1000}, when running on the Kunpeng cluster and utilizing the nondeterministic mode, \FrameworkName{}-SCIP successfully solved 47 instances, with an SGM of 1739.34 and a speedup factor of 22.52, outperforming ParaSCIP by 97.72\%. On the x86 cluster, \FrameworkName{}-SCIP also obtained a significantly better performance than ParaSCIP. The most important techniques that contribute to the better computational performance include:

\begin{enumerate}
    \item Leveraging CP Search in SCIP: The use of CP search techniques in SCIP in the preprocessing and ramp-up phases enables \FrameworkName{}-SCIP to solve certain problem types with high efficiency. 
    \item Adaptive Selection of Workers: The workers are selected and activated adaptively based on the problem features and computing cluster information. This approach enables the successful solution of instances that might otherwise fail due to memory limitations.
    \item Reliable Instance Transferring Method: The presolved instance is exported by the supervisor in the MPS format and transmitted to other machines. As a solver-independent standard for representing MILP problems, the MPS format ensures that the transfer process is reliable, efficient, and straightforward to implement. Moreover, this approach guarantees consistency between the instances processed by the supervisor and those handled by workers, which is critical for correct and reliable distributed solving.
    \item Effective Multi-Presolving: In most cases, presolving takes a short time to reduce the problem size and tighten bounds. Therefore, it is useful to presolve the original problem using multiple strategies and select the smallest presolved problem. This significantly reduces the computational effort in the solving process.
\end{enumerate}

\begin{table}[htbp]
    \centering
    \caption{Results of nondeterministic ParaSCIP and nondeterministic \FrameworkName{}-SCIP with 1,000 processes}
    \label{tab:nondet_1000}
    \begin{tabular}{l l c r r}
    \toprule
    Cluster & Solver & \# solved & SGM & Speedup \\
    \midrule
    
    \multirow{3}{*}{Kunpeng} & SCIP (single thread) & 35 & 39176.15 & — \\
    
    & ParaSCIP & 38 & 3438.70 & 11.39 \\
    
    & \FrameworkName{}-SCIP & 47 & 1739.34 & 22.52 \\
    
    \hline
    
    \multirow{3}{*}{x86} & SCIP (single thread) & 46 & 21469.84 & — \\
    
    & ParaSCIP & 42 & 3515.11 & 6.11 \\
    
    & \FrameworkName{}-SCIP & 47 & 1688.73 & 12.71 \\

    \bottomrule
    \end{tabular}

\end{table}

\begin{table}[htbp]
    \centering
    \caption{Results of deterministic ParaSCIP and deterministic \FrameworkName{}-SCIP with 1,000 processes}
    \label{tab:det_1000}
    \begin{tabular}{l l c r r}
    \toprule
    Cluster & Solver & \# solved & SGM & Speedup \\
    \midrule
    
    \multirow{3}{*}{Kunpeng} & SCIP (single thread) & 4 & 5176.08 & — \\
    
    & ParaSCIP & 2 & 6739.11 & 0.77 \\
    
    & \FrameworkName{}-SCIP & 7 & 2088.06 & 2.48 \\
    
    \hline
    
    \multirow{3}{*}{x86} & SCIP (single thread) & 6 & 3675.97 & — \\
    
    & ParaSCIP & 3 & 6394.68 & 0.57 \\
    
    & \FrameworkName{}-SCIP & 7 & 1336.62 & 2.75 \\
    
    \bottomrule
    \end{tabular}

\end{table}

As shown in Table~\ref{tab:det_1000}, deterministic \FrameworkName{}-SCIP successfully solved 7 instances within the time limit, and achieving speedup factors of 2.48 and 2.75 on Kunpeng and x86 cluster, respectively. Deterministic ParaSCIP failed to accelerate the solving on the test set. The results are consistent with those presented in~\cite{shinano2018fiberscip}.

\subsubsection{Performance with Different Process Numbers}
To conduct a more comprehensive comparison between ParaSCIP and \FrameworkName{}-SCIP, the experiments were extended to solve problems with 100 and 200 processes. In the 100-processes test, a single server was used as a cluster. For the 200-processes test, two servers were connected as a cluster, with MPI assigning 100 processes to each server. 

The results in Tables \ref{tab:nondet_parascip_n2n_scip_diff_processes} and \ref{tab:det_parascip_n2n_diff_processes} show that \FrameworkName{}-SCIP solved more instances and achieved higher speedup than ParaSCIP across different numbers of processes and clusters. It is noteworthy that the deterministic ParaSCIP with 100 processes achieved a speedup factor larger than 1.0. As shown in Table~\ref{tab:det_parascip_n2n_diff_processes}, the average solving time of the deterministic ParaSCIP with 100 processes is 3032.68 and 2576.45 on Kunpeng and x86 cluster, respectively, which are lower than those of the single-threaded SCIP (5176.08 and 3675.97 on Kunpeng and x86 cluster, respectively, as shown in Table~\ref{tab:det_1000}). However, the speedup decreased with an increase in the number of processes. One possible reason is the token mechanism in the deterministic ParaSCIP, which was also discussed in~\cite{shinano2018fiberscip}.

\begin{table}[htbp]
    \centering
    \caption{Results of nondeterministic ParaSCIP and nondeterministic \FrameworkName{}-SCIP with different number of processes}
    \label{tab:nondet_parascip_n2n_scip_diff_processes}
    \begin{tabular}{l r c r r c r r}
    \toprule
    \multirow{2}{*}{Cluster} 
    & \multirow{2}{*}{\# processes} 
    & \multicolumn{3}{c}{ParaSCIP} 
    & \multicolumn{3}{c}{\FrameworkName{}-SCIP}  \\
    \cmidrule(lr){3-5} \cmidrule(lr){6-8}
    & & \# solved & SGM & Speedup & \# solved  & SGM & Speedup \\
    \midrule
    
    \multirow{3}{*}{Kunpeng} & 100  & 41 & 4241.38 & 9.24 & 46 & 2515.56 & 15.57 \\
    & 200  & 38 & 3797.75 & 10.32 & 47 & 2015.56 & 19.44 \\
    & 1000 & 38 & 3438.70 & 11.39 & 47 & 1739.34 & 22.52 \\
    
    \hline
    
    \multirow{3}{*}{x86} & 100  & 43 & 4082.50 & 5.26 & 48 & 2723.54 & 7.88 \\
    & 200  & 42 & 3736.52 & 5.75 & 47 & 1735.05 & 12.37 \\
    & 1000 & 42 & 3515.11 & 6.11 & 47 & 1688.73 & 12.71 \\

    \bottomrule
    \end{tabular}

\end{table}

\begin{table}[htbp]
    \centering
    \caption{Results of deterministic ParaSCIP and deterministic \FrameworkName{}-SCIP with different number of processes}
    \label{tab:det_parascip_n2n_diff_processes}
    \begin{tabular}{l r c r r c r r}
        \toprule
        \multirow{2}{*}{Cluster}
        & \multirow{2}{*}{\# processes} 
        & \multicolumn{3}{c}{ParaSCIP} 
        & \multicolumn{3}{c}{\FrameworkName{}-SCIP}  \\
        \cmidrule(lr){3-5} \cmidrule(lr){6-8}
        & & \# solved & SGM & Speedup & \# solved  & SGM & Speedup \\
        \midrule
        
        \multirow{3}{*}{Kunpeng} & 100  & 5 & 3032.68 & 1.71 & 7 & 2186.15 & 2.37 \\
        & 200  & 6 & 3716.73 & 1.39 & 7 & 2133.78 & 2.43 \\
        & 1000 & 2 & 6739.11 & 0.77 & 7 & 2088.06 & 2.48 \\
        
        \hline
        
        \multirow{3}{*}{x86} & 100  & 6 & 2576.45 & 1.43 & 7 & 1498.73 & 2.45 \\
        & 200  & 6 & 3061.75 & 1.20 & 7 & 1404.09 & 2.62 \\
        & 1000 & 3 & 6394.68 & 0.57 & 7 & 1336.62 & 2.75 \\

        \bottomrule
    \end{tabular}

\end{table}

\subsection{Computational Performance of \FrameworkName{}-HiGHS}

To validate the generality of the \FrameworkName{} framework, another renowned open-source MILP
solver, HiGHS, was also integrated. The performance of \FrameworkName{}-HiGHS was also evaluated with different numbers of processes and on different clusters. When testing the nondeterministic \FrameworkName{}-HiGHS, four instances were removed from the test set because indicator constraints exist in these instances, which HiGHS cannot process.

The summarized results of \FrameworkName{}-HiGHS are presented in Tables \ref{tab:nondet_n2n_highs} and \ref{tab:det_n2n_highs}. Compared with the single-threaded HiGHS, \FrameworkName{}-HiGHS achieved better solving performance. However, the speedup of \FrameworkName{}-HiGHS was lower than that of \FrameworkName{}-SCIP. Possible reasons include: (1) processing the root node in HiGHS is more time-consuming, resulting in a longer ramp-up phase; (2) the average number of tree nodes processed by HiGHS is approximately 25\% of that processed by SCIP, limiting parallel utilization of workers on open nodes; (3) presolving is disabled when HiGHS is used to solve subproblems, as discussed in Section 2.2; and (4) SGM of single-threaded HiGHS is considerably lower than that of single-threaded SCIP, making it inherently more challenging to achieve high speedup.

The results of \FrameworkName{}-HiGHS indicate that achieving substantial speedup with a parallel framework requires certain conditions for the base solver:

\begin{enumerate}
    \item The search trees must contain a sufficient number of nodes to fully utilize parallel computing resources.
    \item The base solver should provide flexible and necessary APIs or callback interfaces to enable the parallel framework to acquire data and implement algorithm.
    \item Parameters for tuning the root node processing strategy are necessary to prevent excessive time consumption at the root node.
\end{enumerate}

\begin{table}[htbp]
    \centering
    \caption{Results of nondeterministic \FrameworkName{}-HiGHS with different number of processes}
    \label{tab:nondet_n2n_highs}
    \begin{tabular}{l l l l r r}
        \toprule
        Cluster & Solver & \# processes & \# solved & SGM & Speedup \\
        \midrule
        
        \multirow{4}{*}{Kunpeng} & HiGHS (single thread) & 1 & 38 & 12762.78 & — \\
        
        & \FrameworkName{}-HiGHS & 100 & 40 & 4977.47 & 2.56 \\
        & \FrameworkName{}-HiGHS & 200 & 40 & 3978.27 & 3.21 \\
        & \FrameworkName{}-HiGHS & 1000 & 42 & 2890.14 & 4.42 \\
        
        \hline
        
        \multirow{4}{*}{x86} & HiGHS (single thread) & 1 & 41 & 8039.96 & — \\
        
        & \FrameworkName{}-HiGHS & 100 & 41 & 4270.42 & 1.88 \\
        & \FrameworkName{}-HiGHS & 200 & 41 & 3505.88 & 2.29 \\
        & \FrameworkName{}-HiGHS & 1000 & 43 & 2560.24 & 3.14 \\

        \bottomrule
    \end{tabular}

\end{table}

\begin{table}[htbp]
    \centering
    \caption{Results of deterministic \FrameworkName{}-HiGHS with different number of processes}
    \label{tab:det_n2n_highs}
    \begin{tabular}{l l l l r r}
        \toprule
        Cluster & Solver & \# processes & \# solved  & SGM & Speedup \\

        \midrule
        
        \multirow{4}{*}{Kunpeng} & HiGHS (single thread) & 1 & 10 & 576.67 & — \\
        
        & \FrameworkName{}-HiGHS & 100 & 10 & 460.90 & 1.25 \\
        & \FrameworkName{}-HiGHS & 200 & 10 & 461.08 & 1.25 \\
        & \FrameworkName{}-HiGHS & 1000 & 10 & 463.38 & 1.24 \\

        \hline
        \multirow{4}{*}{x86} & HiGHS (single thread) & 1 & 9 & 389.76 & — \\
        
        & \FrameworkName{}-HiGHS & 100 & 9 & 282.20  & 1.38 \\
        & \FrameworkName{}-HiGHS & 200 & 9 & 284.04  & 1.37 \\
        & \FrameworkName{}-HiGHS & 1000 & 9 & 292.80  & 1.33 \\

        \bottomrule
    \end{tabular}

\end{table}

\section{Conclusions and Future Directions}\label{sec:Conclusion}

\FrameworkName{}, a parallel framework for solving large-scale MILP problems under distributed memory environment, was  proposed in this study. A novel sliding-window-based deterministic algorithm was designed and implemented in the framework to ensure that the solving tasks are generated and solved in a deterministic order. In addition, advanced techniques have been investigated to leverage parallel computing resources and the solving capabilities of base solvers. Two popular open-source MILP solvers, SCIP and HiGHS, were integrated into the framework, demonstrating that it is generic and extensible. Comprehensive computational experiments were conducted, and the results showed that the framework can accelerate MILP solvers across different modes, numbers of processes and clusters. A speedup factor of 22.52 was achieved by the nondeterministic \FrameworkName{}-SCIP on the Kunpeng cluster, which is significantly higher than that of the state-of-the-art solver ParaSCIP.

Future research directions include: (1) improving the efficiency of the ramp-up phase, that is, generating sufficient open nodes more quickly; (2) investigating more efficient deterministic algorithms that can scale up to large-scale clusters; (3) designing more robust and adaptive load balancing mechanisms; (4) developing novel and diverse solving strategies to leverage parallel computing resources to solve large-scale MILP problems; and (5) combining different types of base solvers into one parallel MILP solver to take advantage of the solving capabilities of different solvers.

%
%

%
%
%
\bibliographystyle{splncs04}
\bibliography{n2n}

\section{Appendix}

\begin{longtable}{l | r r | r r}
  \caption{Solving times (in seconds) of SCIP (single thread) and HiGHS (single thread) on nondeterministic mode test set on two different clusters}
  \label{tab:detail_scip_highs_non_det} \\
  \toprule
  \multirow{2}{*}{Name} & \multicolumn{2}{c|}{SCIP (single thread)} & \multicolumn{2}{c}{HiGHS (single thread)} \\
  \cmidrule{2-5}
  & Kunpeng & x86 & Kunpeng & x86 \\
  \midrule
  \endfirsthead

  \caption{Solving times (in seconds) of SCIP (single thread) and HiGHS (single thread) on nondeterministic mode test set on two different clusters (Continued)}
  \label{tab:detail_scip_highs_non_det:cont} \\
  \toprule
  \multirow{2}{*}{Name} & \multicolumn{2}{c|}{SCIP (single thread)} & \multicolumn{2}{c}{HiGHS (single thread)} \\
  \cmidrule{2-5}
  & Kunpeng & x86 & Kunpeng & x86 \\
  \midrule
  \endhead

  \midrule
  \multicolumn{5}{r}{(To be continued)} \\
  \endfoot

  \bottomrule
  \endlastfoot

    a1c1s1 & 31979.42 & 18016.94 & 6781.61 & 6627.34 \\
    arki001 & 9249.51 & 5055.22 & 2625.44 & 1798.66 \\
    atlanta-ip & 19219.23 & 10176.12 & 17451.12 & 15434.49 \\
    bab5 & 29609.82 & 13070.65 & 734.72 & 450.03 \\
    bley\_xs2 & 39326.74 & 24993.03 & 86400.00 & 86400.00 \\
    blp-ar98 & 35318.73 & 11744.10 & 7694.41 & 3516.84 \\
    blp-ic97 & 52522.95 & 18297.41 & 12182.22 & 6653.35 \\
    blp-ic98 & 64064.30 & 21101.27 & 1886.11 & 1292.90 \\
    bppc8-09 & 25286.65 & 13115.67 & 2712.33 & 2572.93 \\
    cmflsp50-24-8-8 & 55771.47 & 23523.07 & 28805.21 & 13638.00 \\
    comp07-2idx & 86400.00 & 83276.84 & 10711.52 & 3636.99 \\
    cvrpsimple2i & 13174.16 & 9134.23 & - & - \\
    dws008-01 & 86400.00 & 36813.84 & 48936.79 & 39518.02 \\
    eilA101-2 & 86400.00 & 45269.60 & 11114.55 & 5286.67 \\
    gfd-schedulen55f2d50m30k3i & 84916.92 & 32292.80 & - & - \\
    graph20-80-1rand & 45343.52 & 21968.57 & 86400.00 & 86400.00 \\
    ic97\_potential & 36991.52 & 28579.76 & 24211.65 & 16273.04 \\
    icir97\_tension & 7913.35 & 4584.56 & 3636.16 & 1758.64 \\
    markshare\_5\_0 & 37814.85 & 22448.10 & 60176.90 & 55444.63 \\
    msc98-ip & 86400.00 & 77831.64 & 15946.95 & 7789.47 \\
    mspsphard01i & 34009.09 & 20599.94 & - & - \\
    mushroom-best & 14049.14 & 5191.68 & 1229.96 & 533.44 \\
    n3div36 & 30413.46 & 10903.47 & 59464.03 & 25438.79 \\
    neos-1112787 & 36511.79 & 14972.46 & 86400.00 & 12138.35 \\
    neos-1445532 & 86400.00 & 86400.00 & 2.06 & 1.47 \\
    neos-2328163-agri & 12115.59 & 6576.98 & 5770.06 & 8390.79 \\
    neos-2978193-inde & 86400.00 & 36767.69 & 80.73 & 27.94 \\
    neos-3631363-vilnia & 86400.00 & 86400.00 & 86400.00 & 86400.00 \\
    neos-3665875-lesum & 16050.89 & 7635.82 & 24226.53 & 21491.03 \\
    neos-4333464-siret & 64491.74 & 28860.05 & 30914.37 & 15889.97 \\
    neos-4408804-prosna & 53244.61 & 44498.28 & 23850.67 & 5599.49 \\
    neos-4966258-blicks & 19056.70 & 9015.72 & 1161.75 & 362.82 \\
    neos-5093327-huahum & 17507.60 & 8065.60 & 73946.02 & 40742.72 \\
    neos-5107597-kakapo & 14180.18 & 6988.26 & 27015.75 & 30408.77 \\
    neos-5115478-kaveri & 30510.10 & 8711.26 & 22154.69 & 13454.31 \\
    neos-5140963-mincio & 10772.33 & 6084.10 & 7369.62 & 6362.73 \\
    no-ip-65059 & 34132.67 & 16069.34 & 10324.19 & 7804.33 \\
    nu120-pr9 & 46705.60 & 23917.61 & 29429.99 & 18009.39 \\
    opm2-z10-s4 & 86400.00 & 86400.00 & 86400.00 & 71292.90 \\
    pg5\_34 & 11213.31 & 8627.02 & 1113.92 & 580.93 \\
    rococoB10-011000 & 37381.73 & 21711.09 & 19825.83 & 23896.52 \\
    rococoC11-011100 & 86400.00 & 51972.71 & 50839.60 & 39787.58 \\
    seymour & 37716.43 & 21800.22 & 86400.00 & 71292.57 \\
    sing326 & 86400.00 & 58056.70 & 60735.69 & 26910.74 \\
    sorrell3 & 86400.00 & 54758.02 & 86400.00 & 86400.00 \\
    sp97ar & 46189.99 & 17382.61 & 28193.06 & 11794.36 \\
    sp97ic & 86400.00 & 77795.57 & 37695.21 & 13504.62 \\
    splice1k1i & 86400.00 & 77669.94 & - & - \\
    supportcase26 & 86400.00 & 26197.37 & 2164.34 & 1352.37 \\
    supportcase3 & 86400.00 & 86400.00 & 86400.00 & 86400.00 \\
\end{longtable}

\begin{longtable}{l | r r | r r}
  \caption{Solving times (in seconds)  of SCIP (single thread) and HiGHS (single thread) on deterministic mode test set on two different clusters}
  \label{tab:detail_scip_highs_det} \\
  \toprule
  \multirow{2}{*}{Name} & \multicolumn{2}{c|}{SCIP (single thread)} & \multicolumn{2}{c}{HiGHS (single thread)} \\
  \cmidrule{2-5}
  & Kunpeng & x86 & Kunpeng & x86 \\
  \midrule
  \endfirsthead

  \caption{Solving times (in seconds)  of SCIP (single thread) and HiGHS (single thread) on deterministic mode test set on two different clusters (Continued)}
  \label{tab:detail_scip_highs_det:cont} \\
  \toprule
  \multirow{2}{*}{Name} & \multicolumn{2}{c|}{SCIP (single thread)} & \multicolumn{2}{c}{HiGHS (single thread)} \\
  \cmidrule{2-5}
  & Kunpeng & x86 & Kunpeng & x86 \\
  \midrule
  \endhead

  \midrule
  \multicolumn{5}{r}{(To be continued)} \\
  \endfoot

  \bottomrule
  \endlastfoot

    mushroom-best & 7200.00 & 5203.88 & 1234.79 & 534.49 \\
    neos-1456979 & 7200.00 & 4207.99 & 783.21 & 561.93 \\
    neos-957323 & 2372.98 & 1122.00 & 303.12 & 158.15 \\
    neos-960392 & 1500.29 & 803.58 & 29.13 & 18.35 \\
    pg5\_34 & 7200.00 & 7200.00 & 1106.96 & 580.86 \\
    physiciansched6-2 & 4107.69 & 2477.39 & 53.74 & 61.80 \\
    supportcase26 & 7200.00 & 7200.00 & 2183.89 & 1351.35 \\
    supportcase33 & 6750.04 & 3390.34 & 543.42 & 347.85 \\
    traininstance6 & 7200.00 & 7200.00 & 690.33 & 461.24 \\
    trento1 & 7200.00 & 7200.00 & 6584.54 & 7200.00 \\
\end{longtable}

\begin{longtable}{l | r r r | r r r}
  \caption{Solving times (in seconds) of ParaSCIP and \FrameworkName{}-SCIP on nondeterministic mode test set on the Kunpeng cluster}
  \label{tab:detail_parascip_n2n_scip_non_det_kunpeng} \\
  \toprule
  \multirow{2}{*}{Name} & \multicolumn{3}{c|}{ParaSCIP} & \multicolumn{3}{c}{\FrameworkName{}-SCIP} \\
  \cmidrule{2-7}
  & 100 & 200 & 1000 & 100 & 200 & 1000 \\
  \midrule
  \endfirsthead

  \caption{Solving times (in seconds) of ParaSCIP and \FrameworkName{}-SCIP on nondeterministic mode test set on the Kunpeng cluster (Continued)}
  \label{tab:detail_parascip_n2n_scip_non_det_kunpeng:cont} \\
  \toprule
  \multirow{2}{*}{Name} & \multicolumn{3}{c|}{ParaSCIP} & \multicolumn{3}{c}{\FrameworkName{}-SCIP} \\
  \cmidrule{2-7}
  & 100 & 200 & 1000 & 100 & 200 & 1000 \\
  \midrule
  \endhead

  \midrule
  \multicolumn{7}{r}{(To be continued)} \\
  \endfoot

  \bottomrule
  \endlastfoot

    a1c1s1 & 2438.37 & 1352.12 & 1186.27 & 4609.86 & 3902.07 & 1586.07 \\
    arki001 & 1556.41 & 1987.86 & 4588.96 & 2436.70 & 220.69 & 3785.23 \\
    atlanta-ip & 86400.00 & 86400.00 & 86400.00 & 3580.97 & 7158.50 & 17960.86 \\
    bab5 & 711.04 & 675.96 & 606.24 & 1094.62 & 733.88 & 481.99 \\
    bley\_xs2 & 4139.48 & 3079.24 & 2271.48 & 7881.15 & 5628.24 & 1894.26 \\
    blp-ar98 & 6514.81 & 4242.38 & 5963.95 & 6181.32 & 5399.67 & 7157.97 \\
    blp-ic97 & 6742.01 & 14265.14 & 10239.85 & 2671.51 & 2533.78 & 3990.33 \\
    blp-ic98 & 3084.01 & 2249.90 & 2346.15 & 2812.87 & 2397.73 & 2439.22 \\
    bppc8-09 & 315.35 & 116.10 & 50.54 & 346.31 & 275.45 & 327.61 \\
    cmflsp50-24-8-8 & 15574.27 & 8277.12 & 5912.81 & 19005.57 & 10798.97 & 7258.42 \\
    comp07-2idx & 56379.66 & 6646.70 & 4268.92 & 33498.73 & 4990.47 & 7072.02 \\
    cvrpsimple2i & 194.81 & 142.71 & 239.50 & 225.22 & 185.20 & 274.32 \\
    dws008-01 & 4630.68 & 2765.66 & 1612.56 & 7224.03 & 86400.00 & 1443.63 \\
    eilA101-2 & 21711.38 & 10195.86 & 6797.31 & 28719.52 & 22122.31 & 6321.02 \\
    gfd-schedulen55f2d50m30k3i & 86400.00 & 86400.00 & 86400.00 & 28523.06 & 34241.97 & 10866.77 \\
    graph20-80-1rand & 86400.00 & 86400.00 & 86400.00 & 86400.00 & 86400.00 & 86400.00 \\
    ic97\_potential & 220.11 & 121.81 & 137.46 & 220.21 & 200.39 & 222.58 \\
    icir97\_tension & 275.05 & 661.98 & 200.96 & 349.76 & 307.93 & 304.12 \\
    markshare\_5\_0 & 75.99 & 2681.75 & 61.47 & 112.10 & 66.31 & 65.45 \\
    msc98-ip & 5043.98 & 2001.09 & 1269.52 & 2203.35 & 1768.56 & 1128.70 \\
    mspsphard01i & 1131.57 & 1207.79 & 1217.29 & 3664.49 & 2842.79 & 438.04 \\
    mushroom-best & 616.55 & 722.88 & 846.25 & 1060.18 & 1083.49 & 1201.99 \\
    n3div36 & 3205.06 & 2474.70 & 3509.32 & 3394.47 & 2240.65 & 3503.93 \\
    neos-1112787 & 86400.00  & 86400.00 & 86400.00 & 86400.00 & 86400.00 & 86400.00 \\
    neos-1445532 & 86400.00  & 86400.00 & 86400.00 & 3.13 & 3.18 & 3.24 \\
    neos-2328163-agri & 448.37 & 604.38 & 1221.63 & 872.39 & 628.52 & 1466.79 \\
    neos-2978193-inde & 3447.32 & 1340.66 & 1860.18 & 2869.16 & 733.17 & 917.43 \\
    neos-3631363-vilnia & 42938.49 & 23975.57 & 35829.27 & 14865.61 & 200.35 & 212.60 \\
    neos-3665875-lesum & 1595.10 & 1137.53 & 902.04 & 4617.92 & 2964.01 & 1578.01 \\
    neos-4333464-siret & 3118.44 & 4167.33 & 1949.69 & 2602.25 & 2628.51 & 2332.82 \\
    neos-4408804-prosna & 47172.71 & 86400.00 & 35018.21 & 86400.00 & 72868.28 & 52004.64 \\
    neos-4966258-blicks & 86400.00  & 86400.00 & 86400.00 & 2197.12 & 3162.19 & 2813.22 \\
    neos-5093327-huahum & 86400.00  & 86400.00 & 86400.00 & 1382.77 & 1193.96 & 5320.11 \\
    neos-5107597-kakapo & 144.98 & 167.34 & 124.66 & 298.26 & 3107.73 & 124.63 \\
    neos-5115478-kaveri & 122.24 & 262.63 & 390.23 & 5964.77 & 2577.80 & 132.74 \\
    neos-5140963-mincio & 162.27 & 91.92 & 85.60 & 161.10 & 95.30 & 97.46 \\
    no-ip-65059 & 601.80 & 410.804 & 500.14 & 2150.87 & 3528.51 & 1130.08 \\
    nu120-pr9 & 1517.61 & 2165.80 & 2508.59 & 1247.35 & 827.54 & 1760.90 \\
    opm2-z10-s4 & 86400.00  & 86400.00  & 86400.00 & 86400.00 & 51081.21 & 55529.20 \\
    pg5\_34 & 74.39 & 92.92 & 79.39 & 82.27 & 71.50 & 111.66 \\
    rococoB10-011000 & 29868.23 & 1399.17 & 3524.74 & 2084.94 & 1460.12 & 4032.54 \\
    rococoC11-011100 & 8573.75 & 86400.00 & 3412.10 & 2865.04 & 2083.23 & 3674.69 \\
    seymour & 4848.05 & 3497.27 & 1324.45 & 4148.29 & 15016.65 & 1372.47 \\
    sing326 & 45401.21 & 42662.86 & 86400.00 & 50934.36 & 40459.54 & 86400.00 \\
    sorrell3 & 10598.70 & 7430.94 & 86400.00 & 3246.66 & 2468.81 & 9792.11 \\
    sp97ar & 13106.49 & 5595.52 & 7899.34 & 10276.76 & 8219.20 & 18870.11 \\
    sp97ic & 2651.02 & 1764.99 & 4376.78 & 3584.44 & 2474.57 & 4159.14 \\
    splice1k1i & 40871.88 & 86400.00 & 86400.00 & 5566.52 & 5783.77 & 6567.29 \\
    supportcase26 & 32.71 & 25.59 & 41.81 & 24.23 & 33.37 & 31.52 \\
    supportcase3 & 86400.00  & 86400.00 & 86400.00 & 25.00 & 25.06 & 25.40 \\

\end{longtable}

\begin{longtable}{l | r r r | r r r}
  \caption{Solving times (in seconds) of ParaSCIP and \FrameworkName{}-SCIP on nondeterministic mode test set on the x86 cluster}
  \label{tab:detail_parascip_n2n_scip_non_det_x86} \\
  \toprule
  \multirow{2}{*}{Name} & \multicolumn{3}{c|}{ParaSCIP} & \multicolumn{3}{c}{\FrameworkName{}-SCIP} \\
  \cmidrule{2-7}
  & 100 & 200 & 1000 & 100 & 200 & 1000 \\
  \midrule
  \endfirsthead

  \caption{Solving times (in seconds) of ParaSCIP and \FrameworkName{}-SCIP on nondeterministic mode test set on the x86 cluster (Continued)}
  \label{tab:detail_parascip_n2n_scip_non_det_x86_cont} \\
  \toprule
  \multirow{2}{*}{Name} & \multicolumn{3}{c|}{ParaSCIP} & \multicolumn{3}{c}{\FrameworkName{}-SCIP} \\
  \cmidrule{2-7}
  & 100 & 200 & 1000 & 100 & 200 & 1000 \\
  \midrule
  \endhead

  \midrule
  \multicolumn{7}{r}{(To be continued)} \\
  \endfoot

  \bottomrule
  \endlastfoot

    a1c1s1 & 3377.46 & 1963.63 & 1587.31 & 5312.38 & 5078.13 & 1740.77 \\
    arki001 & 269.61 & 1343.03 & 479.85 & 384.91 & 242.11 & 290.64 \\
    atlanta-ip & 86400.00 & 86400.00 & 86400.00 & 3283.62 & 5023.62 & 12435.98 \\
    bab5 & 1790.74 & 1595.28 & 1500.72 & 1817.36 & 1283.26 & 1401.58 \\
    bley\_xs2 & 6542.89 & 7097.42 & 2444.60 & 17184.83 & 8996.50 & 1792.75 \\
    blp-ar98 & 4251.45 & 3049.34 & 8432.82 & 14999.84 & 3826.33 & 4739.20 \\
    blp-ic97 & 3021.08 & 12067.62 & 12793.37 & 2716.50 & 1822.02 & 12363.88 \\
    blp-ic98 & 1441.32 & 2673.53 & 4628.17 & 2335.74 & 1763.18 & 4597.08 \\
    bppc8-09 & 178.80 & 81.01 & 93.48 & 749.79 & 317.98 & 246.77 \\
    cmflsp50-24-8-8 & 6496.56 & 7657.45 & 7460.79 & 8488.33 & 3863.35 & 6074.69 \\
    comp07-2idx & 50317.08 & 9864.82 & 7418.49 & 22306.17 & 6869.32 & 6391.49 \\
    cvrpsimple2i & 171.83 & 178.59 & 336.79 & 248.36 & 396.35 & 304.40 \\
    dws008-01 & 4244.01 & 3350.24 & 1484.81 & 4275.51 & 3423.11 & 1520.39 \\
    eilA101-2 & 18088.52 & 37365.74 & 22157.66 & 23830.67 & 19129.02 & 14508.51 \\
    gfd-schedulen55f2d50m30k3i & 86400.00 & 86400.00 & 86400.00 & 25420.25 & 23753.33 & 12911.88 \\
    graph20-80-1rand & 86400.00 & 86400.00 & 86400.00 & 86400.00 & 86400.00 & 86400.00 \\
    ic97\_potential & 210.32 & 245.09 & 128.04 & 269.50 & 196.87 & 213.73 \\
    icir97\_tension & 393.35 & 279.65 & 366.66 & 496.68 & 423.79 & 175.04 \\
    markshare\_5\_0 & 76.55 & 72.95 & 364.94 & 76.31 & 76.52 & 57.87 \\
    msc98-ip & 5689.76 & 2495.68 & 1236.26 & 10319.81 & 7323.88 & 1226.05 \\
    mspsphard01i & 2007.64 & 2082.68 & 1622.55 & 2212.19 & 365.15 & 354.10 \\
    mushroom-best & 722.81 & 408.57 & 1372.37 & 545.12 & 484.45 & 680.74 \\
    n3div36 & 11234.73 & 4262.77 & 3591.85 & 3874.54 & 3353.99 & 2613.67 \\
    neos-1112787 & 86400.00 & 86400.00 & 86400.00 & 86400.00 & 86400.00 & 86400.00 \\
    neos-1445532 & 86400.00 & 86400.00 & 86400.00 & 4.23 & 4.62 & 5.39 \\
    neos-2328163-agri & 466.07 & 783.60 & 1528.89 & 1027.94 & 812.24 & 1734.24 \\
    neos-2978193-inde & 20420.36 & 1799.34 & 1081.08 & 12776.02 & 1456.92 & 1197.99 \\
    neos-3631363-vilnia & 26666.00 & 14345.33 & 22344.91 & 9523.56 & 257.67 & 235.66 \\
    neos-3665875-lesum & 1940.12 & 1240.34 & 898.05 & 3825.37 & 3581.49 & 2333.24 \\
    neos-4333464-siret & 4390.57 & 2783.28 & 1523.43 & 5449.83 & 3455.01 & 1998.68 \\
    neos-4408804-prosna & 43886.27 & 54221.68 & 35154.87 & 43243.55 & 45610.53 & 59519.91 \\
    neos-4966258-blicks & 86400.00 & 86400.00 & 86400.00 & 4906.89 & 5455.27 & 4999.59 \\
    neos-5093327-huahum & 86400.00 & 86400.00 & 86400.00 & 1327.19 & 1133.74 & 3395.86 \\
    neos-5107597-kakapo & 394.83 & 173.36 & 213.34 & 889.98 & 226.87 & 210.28 \\
    neos-5115478-kaveri & 194.74 & 3656.74 & 304.08 & 5123.21 & 164.86 & 222.71 \\
    neos-5140963-mincio & 154.29 & 101.16 & 89.60 & 160.32 & 82.66 & 98.97 \\
    no-ip-65059 & 447.75 & 380.88 & 462.14 & 2057.14 & 3474.63 & 1015.58 \\
    nu120-pr9 & 1533.73 & 1583.39 & 2761.15 & 1717.10 & 1125.83 & 1734.74 \\
    opm2-z10-s4 & 43874.61 & 45113.93 & 31400.16 & 45028.69 & 86400.00 & 36213.81 \\
    pg5\_34 & 111.74 & 102.77 & 99.06 & 118.82 & 78.92 & 102.47 \\
    rococoB10-011000 & 35693.58 & 1400.51 & 3709.46 & 2477.52 & 1649.80 & 3139.19 \\
    rococoC11-011100 & 7828.80 & 7610.09 & 4053.59 & 3279.67 & 2318.19 & 4097.58 \\
    seymour & 5539.57 & 4320.98 & 1227.71 & 26891.31 & 5112.38 & 1460.69 \\
    sing326 & 47286.65 & 45520.92 & 49347.60 & 39579.30 & 40040.62 & 86400.00 \\
    sorrell3 & 6812.12 & 8163.83 & 18853.11 & 3241.33 & 3655.92 & 8884.96 \\
    sp97ar & 3279.36 & 5112.00 & 5082.96 & 7609.79 & 5609.11 & 6703.05 \\
    sp97ic & 6069.83 & 1811.09 & 4844.94 & 3156.87 & 5650.03 & 3926.38 \\
    splice1k1i & 46732.87 & 86400.00 & 86400.00 & 3842.00 & 3719.49 & 4522.26 \\
    supportcase26 & 53.66 & 49.73 & 56.16 & 37.44 & 28.68 & 38.08 \\
    supportcase3 & 5571.00 & 61066.65 & 23625.15 & 24.43 & 24.26 & 25.24 \\

\end{longtable}

\begin{longtable}{l | r r r | r r r}
  \caption{Solving times (in seconds) of \FrameworkName{}-HiGHS on nondeterministic mode test set}
  \label{tab:detail_disthighs_non_det} \\
  \toprule
  \multirow{2}{*}{Name} & \multicolumn{3}{c|}{Kunpeng-\FrameworkName{}-HiGHS}  & \multicolumn{3}{c}{x86-\FrameworkName{}-HiGHS} \\
  \cmidrule{2-7}
  & 100 & 200 & 1000 & 100 & 200 & 1000 \\
  \midrule
  \endfirsthead

  \caption{Solving times (in seconds) of \FrameworkName{}-HiGHS on nondeterministic mode test set (Continued)}
  \label{tab:detail_disthighs_non_det:cont} \\
  \toprule
  \multirow{2}{*}{Name} & \multicolumn{3}{c|}{Kunpeng-\FrameworkName{}-HiGHS}  & \multicolumn{3}{c}{x86-\FrameworkName{}-HiGHS} \\
  \cmidrule{2-7}
  & 100 & 200 & 1000 & 100 & 200 & 1000 \\
  \midrule
  \endhead

  \midrule
  \multicolumn{7}{r}{(To be continued)} \\
  \endfoot

  \bottomrule
  \endlastfoot

    a1c1s1 & 1287.25 & 1003.42 & 760.58 & 1471.43 & 1244.65 & 916.71 \\
    arki001 & 287.20 & 300.39 & 475.61 & 326.43 & 539.63 & 327.52 \\
    atlanta-ip & 15198.38 & 12616.52 & 27691.45 & 14413.98 & 9499.94 & 7792.09 \\
    bab5 & 6951.59 & 1413.33 & 1269.52 & 3884.53 & 4258.69 & 592.73 \\
    bley\_xs2 & 8156.95 & 6764.13 & 4755.19 & 8011.47 & 8159.69 & 6127.94 \\
    blp-ar98 & 6521.69 & 6557.48 & 5896.99 & 7843.28 & 5767.59 & 5522.35 \\
    blp-ic97 & 10496.19 & 6700.27 & 5624.41 & 8159.03 & 5207.81 & 4189.76 \\
    blp-ic98 & 4744.86 & 3331.11 & 3750.57 & 2534.56 & 2343.96 & 1700.37 \\
    bppc8-09 & 206.18 & 221.36 & 162.49 & 179.24 & 258.50 & 201.86 \\
    cmflsp50-24-8-8 & 21974.68 & 13060.42 & 9348.19 & 13382.73 & 8599.35 & 6201.51 \\
    comp07-2idx & 22012.15 & 8964.64 & 5695.73 & 9741.93 & 12267.45 & 4307.10 \\
    dws008-01 & 86400.00 & 86400.00 & 1531.57 & 2315.29 & 1639.00 & 1541.74 \\
    eilA101-2 & 10359.48 & 10539.75 & 10352.08 & 10031.53 & 10435.28 & 10847.68 \\
    graph20-80-1rand & 86400.00 & 86400.00 & 86400.00 & 86400.00 & 86400.00 & 86400.00 \\
    ic97\_potential & 742.14 & 540.34 & 426.47 & 786.32 & 446.51 & 488.83 \\
    icir97\_tension & 998.35 & 872.55 & 764.75 & 1237.64 & 1334.84 & 870.20 \\
    markshare\_5\_0 & 1455.19 & 695.33 & 212.39 & 795.16 & 423.41 & 167.58 \\
    msc98-ip & 20424.16 & 18798.29 & 7712.96 & 11048.45 & 11363.98 & 5697.26 \\
    mushroom-best & 1692.07 & 1285.28 & 1196.39 & 1142.66 & 912.02 & 960.82 \\
    n3div36 & 10284.49 & 9694.73 & 9293.25 & 6814.62 & 6595.30 & 5448.69 \\
    neos-1112787 & 86400.00 & 86400.00 & 86400.00 & 86400.00 & 86400.00 & 86400.00 \\
    neos-1445532 & 3.26 & 3.47 & 5.64 & 4.50 & 4.74 & 7.06 \\
    neos-2328163-agri & 5912.90 & 5775.08 & 6488.15 & 8424.37 & 6539.66 & 14446.47 \\
    neos-2978193-inde & 50.64 & 49.91 & 72.16 & 48.19 & 48.50 & 51.79 \\
    neos-3631363-vilnia & 86400.00 & 86400.00 & 86400.00 & 86400.00 & 86400.00 & 86400.00 \\
    neos-3665875-lesum & 2350.41 & 1999.55 & 1230.67 & 3449.88 & 2280.79 & 1618.00 \\
    neos-4333464-siret & 23900.41 & 11126.49 & 5188.41 & 27122.77 & 11523.38 & 5107.79 \\
    neos-4408804-prosna & 50385.03 & 21123.09 & 6433.01 & 36945.56 & 19400.02 & 4426.15 \\
    neos-4966258-blicks & 1597.26 & 1436.33 & 890.82 & 895.89 & 820.51 & 817.47 \\
    neos-5093327-huahum & 7750.18 & 6209.04 & 6404.06 & 7203.41 & 5306.67 & 5398.72 \\
    neos-5107597-kakapo & 1608.38 & 1646.53 & 1115.68 & 1939.75 & 1365.74 & 1631.17 \\
    neos-5115478-kaveri & 1403.33 & 1716.54 & 816.64 & 2255.70 & 1226.86 & 1150.03 \\
    neos-5140963-mincio & 126.07 & 77.24 & 64.38 & 150.12 & 110.34 & 79.51 \\
    no-ip-65059 & 3561.55 & 3290.81 & 1294.78 & 3816.10 & 2832.94 & 1262.11 \\
    nu120-pr9 & 6570.88 & 6487.00 & 5841.15 & 7457.43 & 7460.66 & 17479.74 \\
    opm2-z10-s4 & 86400.00 & 86400.00 & 86400.00 & 86400.00 & 86400.00 & 65757.40 \\
    pg5\_34 & 125.15 & 123.35 & 135.52 & 234.35 & 169.54 & 185.69 \\
    rococoB10-011000 & 10912.12 & 6406.97 & 3581.05 & 14559.95 & 7128.52 & 2881.24 \\
    rococoC11-011100 & 6313.90 & 3710.83 & 2783.89 & 5438.07 & 4388.34 & 5258.65 \\
    seymour & 60125.87 & 36905.60 & 15294.68 & 45985.60 & 29993.04 & 14878.57 \\
    sing326 & 53059.93 & 51137.03 & 42891.61 & 45189.25 & 44305.38 & 30892.35 \\
    sorrell3 & 9244.35 & 10344.31 & 9870.84 & 15804.73 & 11527.77 & 7928.98 \\
    sp97ar & 18741.46 & 10746.87 & 11199.96 & 9180.61 & 7225.44 & 6576.47 \\
    sp97ic & 6834.12 & 6667.17 & 4091.61 & 5557.16 & 3971.21 & 3492.62 \\
    supportcase26 & 138.09 & 157.28 & 161.43 & 138.14 & 124.32 & 208.26 \\
    supportcase3 & 86400.00 & 86400.00 & 55823.63 & 86400.00 & 86400.00 & 4001.61 \\

\end{longtable}

\begin{longtable}{l | r r r | r r r}
  \caption{Solving times (in seconds) of ParaSCIP and \FrameworkName{}-SCIP on deterministic mode test set on the Kunpeng cluster}
  \label{tab:detail_parascip_n2n_scip_det_kunpeng} \\
  \toprule
  \multirow{2}{*}{Name} & \multicolumn{3}{c|}{ParaSCIP} & \multicolumn{3}{c}{\FrameworkName{}-SCIP} \\
  \cmidrule{2-7}
  & 100 & 200 & 1000 & 100 & 200 & 1000 \\
  \midrule
  \endfirsthead

  \caption{Solving times (in seconds) of ParaSCIP and \FrameworkName{}-SCIP on deterministic mode test set on the Kunpeng cluster (Continued)}
  \label{tab:detail_parascip_n2n_scip_det_kunpeng:cont} \\
  \toprule
  \multirow{2}{*}{Name} & \multicolumn{3}{c|}{ParaSCIP} & \multicolumn{3}{c}{\FrameworkName{}-SCIP} \\
  \cmidrule{2-7}
  & 100 & 200 & 1000 & 100 & 200 & 1000 \\
  \midrule
  \endhead

  \midrule
  \multicolumn{7}{r}{(To be continued)} \\
  \endfoot

  \bottomrule
  \endlastfoot

    mushroom-best & 7200.00 & 7200.00 & 7200.00 & 2503.6 & 2443.8 & 2416.8 \\
    neos-1456979 & 7200.00 & 7200.00 & 7200.00 & 7200.0 & 7200.0 & 7200.0 \\
    neos-957323 & 1228.55 & 2599.76 & 7200.00 & 2409.1 & 2257.9 & 2002.1 \\
    neos-960392 & 1675.75 & 4611.35 & 7200.00 & 1267.7 & 1263.2 & 1273.0 \\
    pg5\_34 & 7200.00 & 7200.00 & 7200.00 & 2568.2 & 2385.1 & 2324.2 \\
    physiciansched6-2 & 263.81 & 559.56 & 4960.49 & 203.1 & 202.8 & 202.1 \\
    supportcase26 & 1804.70 & 2393.06 & 7200.00 & 793.8 & 787.6 & 746.7 \\
    supportcase33 & 3354.53 & 1763.82 & 5392.38 & 2019.5 & 1889.1 & 1868.7 \\
    traininstance6 & 7200.00 & 6520.62 & 7200.00 & 7200.0 & 7200.0 & 7200.0 \\
    trento1 & 7200.00 & 7200.00 & 7200.00 & 7200.0 & 7200.0 & 7200.0 \\

\end{longtable}

\begin{longtable}{l | r r r | r r r}
  \caption{Solving times (in seconds) of ParaSCIP and \FrameworkName{}-SCIP on deterministic mode test set on the x86 cluster}
  \label{tab:detail_parascip_n2n_scip_det_x86} \\
  \toprule
  \multirow{2}{*}{Name} & \multicolumn{3}{c|}{ParaSCIP} & \multicolumn{3}{c}{\FrameworkName{}-SCIP} \\
  \cmidrule{2-7}
  & 100 & 200 & 1000 & 100 & 200 & 1000 \\
  \midrule
  \endfirsthead

  \caption{Solving times (in seconds) of ParaSCIP and \FrameworkName{}-SCIP on deterministic mode test set on the x86 cluster (Continued)}
  \label{tab:detail_parascip_n2n_scip_det_x86:cont} \\
  \toprule
  \multirow{2}{*}{Name} & \multicolumn{3}{c|}{ParaSCIP} & \multicolumn{3}{c}{\FrameworkName{}-SCIP} \\
  \cmidrule{2-7}
  & 100 & 200 & 1000 & 100 & 200 & 1000 \\
  \midrule
  \endhead

  \midrule
  \multicolumn{7}{r}{(To be continued)} \\
  \endfoot

  \bottomrule
  \endlastfoot

    mushroom-best & 7200.00 & 7200.00 & 7200.00 & 1233.0 & 1113.8 & 1026.2 \\
    neos-1456979 & 6635.28 & 7200.00 & 7200.00 & 7200.0 & 7200.0 & 7200.0 \\
    neos-957323 & 920.96 & 1947.51 & 7200.00 & 1274.7 & 1081.1 & 941.6 \\
    neos-960392 & 1544.43 & 4162.55 & 7200.00 & 677.4 & 683.5 & 689.0 \\
    pg5\_34 & 7200.00 & 7200.00 & 7200.00 & 1902.4 & 1553.6 & 1375.2 \\
    physiciansched6-2 & 180.04 & 402.41 & 3998.68 & 118.6 & 119.2 & 120.1 \\
    supportcase26 & 882.19 & 1237.01 & 6714.55 & 540.5 & 481.2 & 443.3 \\
    supportcase33 & 3030.70 & 1509.01 & 4243.68 & 1087.1 & 1000.9 & 919.2 \\
    traininstance6 & 7200.00 & 4327.68 & 7200.00 & 7200.0 & 7200.0 & 7200.0 \\
    trento1 & 7200.00 & 7200.00 & 7200.00 & 7200.0 & 7200.0 & 7200.0 \\

\end{longtable}

\begin{longtable}{l | r r r | r r r}
  \caption{Solving times (in seconds) of \FrameworkName{}-HiGHS on deterministic mode test set}
  \label{tab:detail_n2n_highs_det} \\
  \toprule
  \multirow{2}{*}{Name} & \multicolumn{3}{c|}{Kunpeng-\FrameworkName{}-HiGHS} & \multicolumn{3}{c}{x86-\FrameworkName{}-HiGHS} \\
  \cmidrule{2-7}
  & 100 & 200 & 1000 & 100 & 200 & 1000 \\
  \midrule
  \endfirsthead

  \caption{Solving times (in seconds) of \FrameworkName{}-HiGHS on deterministic mode test set (Continued)}
  \label{tab:detail_n2n_highs_det:cont} \\
  \toprule
  \multirow{2}{*}{Name} & \multicolumn{3}{c|}{Kunpeng-\FrameworkName{}-HiGHS} & \multicolumn{3}{c}{x86-\FrameworkName{}-HiGHS} \\
  \cmidrule{2-7}
  & 100 & 200 & 1000 & 100 & 200 & 1000 \\
  \midrule
  \endhead

  \midrule
  \multicolumn{7}{r}{(To be continued)} \\
  \endfoot

  \bottomrule
  \endlastfoot

    mushroom-best & 1499.9 & 1489.7 & 1510.2  & 535.8 & 528.3 & 535.8 \\
    neos-1456979 & 925.3 & 923.2 & 921.1  & 279.9 & 279.2 & 285.2 \\
    neos-957323 & 208.7 & 209.5 & 211.9  & 109.3 & 117.6 & 124.8  \\
    neos-960392 & 30.2 & 30.4 & 32.3  & 30.4 & 30.7 & 34.8 \\
    pg5\_34 & 484.8 & 480.7 & 479.6  & 189.9 & 185.4 & 185.7 \\
    physiciansched6-2 & 123.3 & 122.0 & 122.7  & 44.2 & 44.7 & 49.4  \\
    supportcase26 & 323.6 & 328.3 & 322.4  & 695.6 & 689.7 & 698.9  \\
    supportcase33 & 626.1 & 643.5 & 636.2 & 217.6 & 220.8 & 223.6 \\
    traininstance6 & 873.0 & 866.9 & 860.0  & 464.2 & 470.2 & 469.8 \\
    trento1 & 3483.8 & 3443.8 & 3493.4 & 7200.0 & 7200.0 & 7200.0  \\

\end{longtable}

\end{document}